\documentclass[letterpaper, 10 pt, conference]{ieeeconf}  
\pdfoutput=1

\IEEEoverridecommandlockouts                              

\usepackage[noadjust]{cite}
\usepackage{graphicx}
\usepackage{xcolor}
\usepackage{booktabs}
\usepackage{url}

\title{\LARGE \bf
Object Recognition from Short Videos for Robotic Perception
}

\author{Ivan Bogun$^{1}$, Anelia Angelova$^{2}$ and Navdeep Jaitly$^{2}$
\thanks{*This work was done while the author was with Google Research.}
\thanks{$^{1}$Ivan Bogun is with the Department of Computer Science \& Cybersecurity,
Florida Institute of Technology, Melbourne, Florida, USA
        {\tt\small ibogun2010@my.fit.edu}}%
\thanks{$^{2}$Anelia Angelova and Navdeep Jaitly are with Google Research.
        {\tt\small anelia@google.com}}%
}
\DeclareGraphicsExtensions{.pdf,.png,.jpg}
\begin{document}

\maketitle
\thispagestyle{empty}
\pagestyle{empty}

\begin{abstract}
Deep neural networks have become the primary learning technique for object
recognition. Videos, unlike still images,
are temporally coherent which makes the application of deep networks non-trivial. Here,
we investigate how  motion can aid object recognition in short videos. Our approach
is based on Long Short-Term Memory (LSTM) deep networks. Unlike previous applications
of LSTMs, we implement each gate as a convolution. We show that
convolutional-based LSTM models are capable of learning
motion dependencies and are able to improve the recognition accuracy when more frames in a sequence are available.
We evaluate our approach on the Washington RGBD Object dataset and on the Washington
RGBD Scenes dataset. Our approach outperforms deep nets applied to still images and sets
a new state-of-the-art in this domain.
\end{abstract}

\section{INTRODUCTION}

Deep neural networks (DNNs) have been established as a predominant method for object
recognition. 
 By taking advantage of large datasets and learning capacity, deep
 networks are able  learn to recognize many object categories. Generalizing deep
networks to work on  videos is a hard problem since video frames are highly correlated. Here, we study  the problem of object  recognition from  short videos (up to 5 frames). This is a common scenario in robotics perception, for example,
 a camera-mounted robotic arm manipulator can record
a small video as it approaches an object, and use it for better recognition.  Similarly, when data is acquired by a mobile phone, a short video sequence can be taken instead of still image for better recognition. We argue that the motion, readily available in the videos, should
be used as an additional cue for recognition and present a  method capable of
taking advantage of it. Our method is based on  recurrent convolutional neural network where we use
convolutional Long Short-Term Memory (LSTM) layer to capture motion information.
\par We summarize our contributions as follows:
\begin{itemize}
\item Introduce a fast baseline DNN network which achieves accuracy competitive with the state-of-the-art on the Washinton-RGBD Object dataset, while running at 0.1 seconds per image on GPU and without using segmentation masks.
\item Present a motion model, based on LSTM, which uses convolution at each of
its gates, and show that it is advantageous for recognition and can generalize to more frames. The model
uses fewer parameters than the baseline and runs at 0.87 seconds per two frame sequences, which makes it practically relevant. To our knowledge this is the first application of fully-convolutional LSTM networks.
\item Experiment extensively on Washington-Object and Washington-Scenes datasets and set a new
state-of-the-art in the domain on both datasets.
\end{itemize}

\section{Previous work}

Object recognition for robotics applications has been of interest for a number of
 years~\cite{gould2008integrating} with more advanced and sophisticated algorithms \cite{bo2013unsupervised,lai2011large} making their
 way into this application domain~\cite{jia2011robotic, collet2009object, pillai2015monocular}. Deep neural networks have also recently
 been applied to the problems, such as object recognition and grasp detection, and demonstrated advancements \cite{shwarz2015object, lenz2015deep}.

LSTMs have been introduced by Hochreiter and Schmidhuber with their
pioneering work in~\cite{Hochreiter97}.
A. Graves and collaborators~\cite{Graves08,Graves14ICML} further developed
LSTMs and showed important improvements and practical applications to
various domains with sequence data-processing, such as handwriting
recognition~\cite{Graves08,Graves09} and speech recognition~\cite{Graves13,Graves14ICML}.
Recurrent neural networks have also recently been successfully applied for phoneme
 recognition~\cite{Fernandez08}, translation~\cite{Suteskever14} or for generating
 text descriptions of input images~\cite{vinyals2014show}.
 \par In video analysis, several applications of RNNs or LSTMs have been proposed for
 improving recognition from consecutive image frames. For example,
Karpathy et al.~\cite{Karpathy14} applied a 3D convolutions for classification of
 sports videos, and Ng et al.~\cite{Ng15} applied a combination of LSTMs and global
 pooling techniques for classification of sports videos, or other
 action recognition videos.
The key difference with our work is that instead of LSTM, based on fully connected gates, we use convolutional-only LSTM gates
which allow us to model local motion deformations. Additionally, other approaches are typically applied for long video
 sequences. Because of high computational cost, they have to be applied offline and can afford much larger and slower
 LSTM architectures, or may choose to combine LSTM-based methods with optical flow,
or others techniques~\cite{Ng15}.  We, on the other hand, focus on object recognition from small sequences of
image frames, e.g. taken within one second, which is relevant to robotics perception or mobile phone recognition.

\section{Approach}
\label{sec:Method}

\subsection{Baseline convolutional neural network}
\label{sub:baseline_conv_network}
Since winning ImageNet challenge for
image recognition \cite{Krizhevsky12}, convolutional neural networks
(CNNs)  have become widely used in object recognition. By stacking multiple layers on top of
each other, CNNs can learn progressively higher level representations of the
images thus performing feature learning. For image-based inputs, the first layers are typically convolutional, targeted at learning local visual features, whereas the subsequent layers are fully-connected.
\par  We first build
a baseline CNN architecture for object recognition for a single frame. Inspired by \cite{Krizhevsky12}, we build our baseline model using three convolutional layers, followed by two fully connected layers with dropouts. We note that unlike~\cite{Krizhevsky12}, our network has been designed to be very small (fewer and smaller convolutional filters), and thus is very fast at inference. 
The supplementary material provides  details of the architecture of this network and visualizes it.

\subsection{Long Short-Term memory}
\label{sub:long_short_term_memory}
 Recurrent neural networks (RNNs) represent
a more general class of networks, where connections from hidden units can be used
as input to the network~\cite{Hochreiter97,Graves08}. Because of recurrent connections, RNNs are capable of
learning dynamic relationship in the
data. One of drawbacks of RNNs is that during training, which is done by backpropagation, the gradient may either
become very small or too large, problems known as ``vanishing gradient'' and
 ``exploding gradient'',  respectively. To solve this problem, Long Short-Term
 Memory (LSTM) \cite{hochreiter1997long} introduced special \emph{memory cells}
 to store information for multiple time steps.
LSTM architecture is a block consisting of an input gate, a neuron which has a
recurrent connection, a forget gate and an output gate. The motivation behind such
blocks is to allow the network to read (input gate), write (output gate)  and
reset (forget gate), similar to operations available to a digital computer. Formally, the
transformation taking place in an LSTM layer is as follows: let $\{x_t\}$ be an
input sequence where $t$ is a timestep. The input gate, $i_t$, and an internal
gate which is a
candidate to be placed into a memory block, $\hat{C_t}$, are:
\begin{equation}
  i_t = \sigma (W_i x_t + U_i h_{t-1} + b_i)
\end{equation}
\begin{equation}
  \hat{C_t} = tanh(W_c x_t + U_c h_{t-1} +b_c)
\end{equation}
where $h_{t-1}$ is the LSTM output at timestep $t-1$ and $W_i, U_i, b_i, W_c, U_c, b_c$ are the weight
matrices and biases for the input gate and the memory block, and $\sigma$ is the sigmoid function. Note that all gates are a function
of the input at a current time step, $x_t$, and the output of previous step,
$h_{t-1}$, forming recurrent dependency. In the original LSTM design
\cite{hochreiter1997long}, the
content of the memory cell at the time step $t$ is a linear combination of the
current memory cell candidate and the content of the cell's memory at previous
time step:
\begin{equation}
  C_t = i_t * \hat{C_t} + C_{t-1}
\end{equation}
where $*$ is the element-wise addition operation. This could lead to
unbounded values of the memory cell. A solution is to reset some of the
elements of the previous memory cell with the aid of the forget gate \cite{gers2000learning}.
Empirically it was  found that the forget gate plays a crucial role in the LSTM
design and removing it decreases performance \cite{greff2015lstm}.
The forget gate, $f_t$ and the memory state, $C_t$ are computed as follows:
\begin{equation}
  f_t  = \sigma(W_f x_t +U_f h_{t-1}+ b_f).
\end{equation}
\begin{equation}
  C_t = i_t * \hat{C_t} + f_t *C_{t-1}.
\end{equation}
The forget gate serves as a way to reset activations from previous memory cell
which might not be as important at the current time step. Note that all this
behavior is learned via the weights of the LSTM and is thus adapted to the data
observed. The final output is given as a function of the output gate multiplied by the
activation of the memory cell at the current time step:
\begin{equation}
  o_t = \sigma(W_o x_t + U_o h_{t-1} + V_o C_t + b_o)
\end{equation}
\begin{equation}
  h_t = o_t * tanh(C_t)
\end{equation}
Because of memory cells, LSTM can transfer information across different timesteps,
an essential feature if we are to use dynamics of the input to aid recognition.

\subsection{Convolutional LSTM for modeling motion}
\label{sub:Conv_vs_fc_lstm}
Not all motion information in sequences of frames is useful when recognizing objects.
For example, camera motion or the motion of the environment (e.g., turning motion
of the turntable with the object of the interest) is not relevant to the identity
of the object. Thus, the neural network's architecture has to be designed in a way as to
account for the situation in which the relevant motion is local. For example, when a
cup rotates on a turntable, only the shape of the handle deforms, while the overall
shape of the cup remains the same. To extract such local motion information, we propose to
implement the LSTM module in terms of convolutions, instead of fully connected layers, as was previously done. Since
each gate in the LSTM is convolutional, the recurrent network is capable to act upon local motion
from the video which is specific to each object. Another benefit of convolutional-only based
 LSTM is that it requires significantly fewer
 parameters, compared to a fully connected LSTM. This allows to reduce overfitting, and generalizes better, which is particularly important
especially when the training data is limited. Our experiments, presented later in the paper, further validate that the proposed fully-convolutional LSTM is a much better architecture in terms of accuracy for this task, compared to other alternatives.

\subsection{Bidirectional LSTM}
\label{sub:bidirectional_lstm}
\par  The LSTM architecture, presented in Sec. \ref{sub:long_short_term_memory}, allows
dynamic information to pass only in one direction, i.e. for timesteps $t_i, t_j$, such that $t_i < t_j $, the output of $h_{t_j}$ has
 access to the information available at previous timestep $h_{t_i}$, but $h_{t_i}$ does not have access to $h_{t_j}$. A bidirectional LSTM architecture was
proposed \cite{graves2005framewise} to overcome this. In
bidirectional LSTM, the input sequence is passed into two separate LSTM layers in forward 
and reverse directions.
 In Sec. \ref{sub:finding_best_train_test}
we further explore how presenting the input sequences in one or both directions
affect the network's learning capabilities.

\subsection{Motion model for image sequences}
\label{sub:network_architecture}

This section summarizes the architecture of our proposed convolutional LSTM-based motion model.

The first three layers in our LSTM-based motion model
are convolutional layers, as in the baseline (Section~\ref{sub:baseline_section_details} provides specific details of their parameters). This is done to allow the
network to learn low-level features. These layers are followed by bidirectional LSTM modules with convolutional gates. Each gate of the LSTM is implemented as a convolutional layer with stride 1, filter size 5 and depth 256. The hidden layers from the forward and backward LSTM are concatenated
and fed
into a fully connected layer followed by a softmax layer. The
network architecture is shown in Fig. \ref{fig:bidirectional} and is visualized in more detail in the supplementary material. For 4-frame sequence our motion model uses 66 million parameters whereas the baseline uses 77 million.
As previously mentioned, the convolutional-only and the bidirectional design is proposed to address the problem of learning from short video sequences.

\begin{figure}[h!]
 
  \centering
  \includegraphics[width=0.35\textwidth]{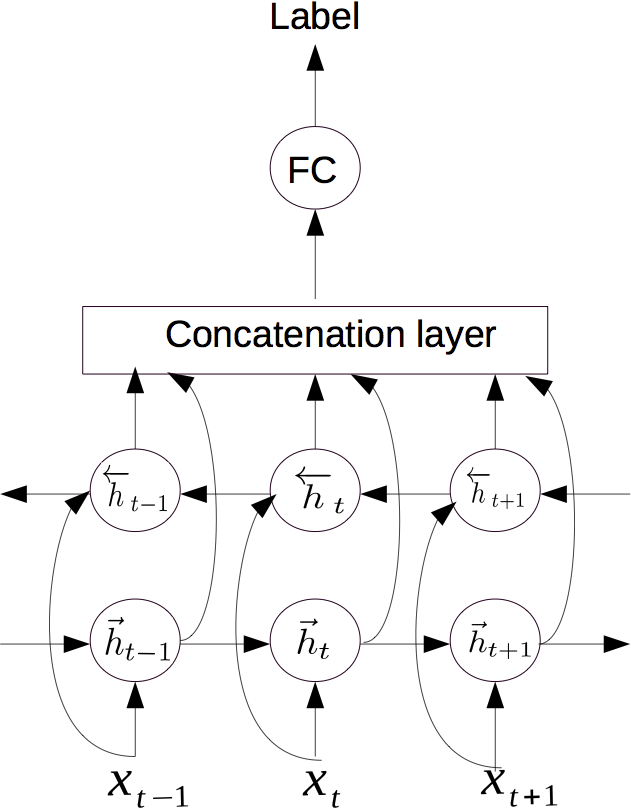}
  \caption{Bidirectional network of LSTMs. The frames are processed with a CNN to obtain a sequence  $\{x_{t-1}, x_t , x_{t+1} \} $ which serves as an input to two LSTM layers.
The results of the forward and backward passes are concatenated and a fully connected layers is put on top. Resulting
class label is found by applying a softmax function.}
 \label{fig:bidirectional}
\end{figure}

\section{Experimental evaluation}
\label{sec:Experiments}
We evaluated our method on publicly available datasets and compared it to the state-of-the-art. First, we tested on Washington RGBD Object dataset \cite{lai2011large}, which has been the most common to test object recognition for robotic perception. Although Washington RGBD Object dataset is one of the largest and most popular, it was recorded in a controlled environment. To evaluate our method in more realistic settings, we used the Washington Scenes dataset\footnote{https://goo.gl/wOAsla} \cite{lai2011large}. The Scenes dataset contains videos recorded in settings, such as a kitchen or an office, with multiple objects occurring naturally.
\subsection{Washington Object dataset results}
\label{sub:washington_object_dataset}   

\begin{figure}[h!]  
  \centering
    \includegraphics[width=0.47\textwidth]{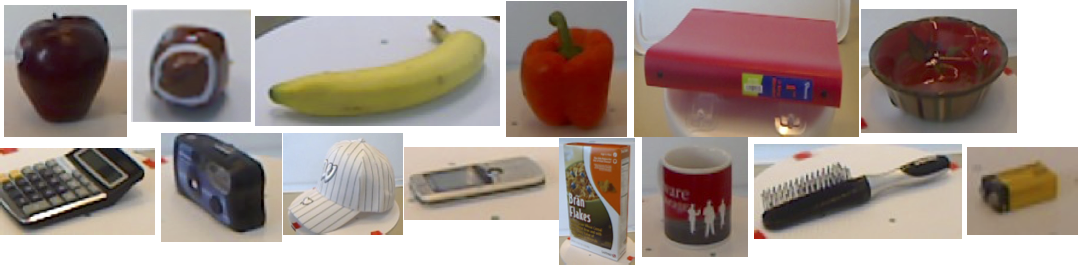}
  \caption{Sample classes from the Washington-RGBD object dataset.}
  \label{fig:dataset} 
\end{figure}

The Washington Object dataset
\cite{lai2011large}  is a collection of
images of objects from $51$ different categories. The dataset was collected by taking
images of the objects located on a turntable from 3 different viewing
angles. The dataset contains around 250,000 images from objects such as flashlight,
cup, ball, apple and others. From all the images only every fifth frame is used for
training or testing. Fig. \ref{fig:dataset} contains a sample set of classes in
the dataset. The standard evaluation protocol \cite{lai2011large} uses $10$ cross-validation
splits when reporting results, which we also follow in this paper. It is worth noting
 that the dataset contains extra information e.g. segmentation masks and depth images
which are not used in our experiments neither for training nor for testing.

\subsubsection{Summary of results}

\par
Table \ref{table:baseline} shows the results of our algorithm on the Washington Object dataset. We present the results of our baseline (Section \ref{sub:baseline_section}) and of our motion model (Section~\ref{sub:network_architecture}). Our motion model is tested in two scenarios: natural (or ``short-time frame'') and ``wide viewpoint''
settings, and for different number of frames. In
the natural setting, the sequences we used are taken less then a second apart - a
common scenario when using mobile device to take a picture of an object in the real world.
In the second scenario the images of the objects are taken as far apart as possible
as to maximize the viewpoints of the object.

\par From our results with motion from short video sequences, we can see that the motion models, in all settings, outperform previous strong baselines on still images. This indicates that motion can be used to enhance performance and with it, a new state-of-the-art for RGB input can be established.
On the wide viewpoint dataset, the motion model outperforms all prior state-of-the-art results and shows a clear pattern of generalization when presented with longer sequences.
Additionally, we also compare to a motion-based baseline which uses the same  video sequences but does not apply convolutional LSTM. This method, also shows benefits of the motion information; it can be practically utilized to solve the same problem with somewhat inferior performance to an LSTM-based model. The sections below provide details of these experiments.
\subsubsection{Baseline}
\label{sub:baseline_section}
\par
 Our baseline model (Section \ref{sub:network_architecture}) achieves an accuracy of $82.02 \pm 1.96 \%$ which makes it comparable to the state of the art on the dataset. The method \cite{shwarz2015object} uses additional segmentation masks and has a better overall performance at $83.1 \pm 2$. It is worth noting that our baseline model is pretty fast as it takes $0.1$ second per image to perform recognition.

\begin{table}[h!]
  
  \begin{tabular}{ | p{4cm}| c | c  |}
   \hline Model & Accuracy & Standard Deviation  \\
\toprule
\multicolumn{3}{c}{Single frame dataset.} \\
   \hline Lai \emph{et. al.} ~\cite{lai2011large} & 74.3 & 3.3 \\
   \hline Bosch \emph{et. al.} \cite{bosch2007image} & 80.2 & 1.8 \\
   \hline Socher \emph{et. al.} \cite{socher2012convolutional} & 80.8 & 4.2 \\
    \hline  Bo \emph{et. al.} \cite{bo2013unsupervised} & 82.4 & 3.1  \\
      \hline  Schwarz \emph{et. al.} \cite{shwarz2015object} & \textbf{83.1} & 2  \\
         \hline  Ours: baseline & 82.02 & 1.96 \\
         \hline
         \multicolumn{3}{c}{Short time-frame sequence dataset} \\
          \hline Motion model & \textbf{82.74} & 1.76  \\
           \hline Baseline + multi-frame average pooling during test &  82.66 & 1.8 \\
         \hline
         \multicolumn{3}{c}{Wide viewpoint: $2$ frame dataset} \\
            \hline Motion model& \textbf{83.32} & 1.96  \\
            \hline Baseline + multi-frame average pooling during test &  83.07 & 1.8 \\
           \hline
             \multicolumn{3}{c}{Wide viewpoint: $3$ frame dataset} \\
            \hline Motion model& \textbf{84.29} & 2  \\
              \hline Baseline + multi-frame average pooling during test &  82.45 & 1.9 \\
             \hline
         \multicolumn{3}{c}{Wide viewpoint: $4$ frame dataset} \\
            \hline Motion model& \textbf{84.23} & 1.84  \\
            \hline Baseline + multi-frame average pooling during test &  84.18 & 1.67 \\

     \hline
   \end{tabular}
   \caption{Results of the recognition task on Washington-RGBD object dataset.
   The motion model used unidirectional training and bidirectional testing as
   explained in Sec. \ref{sub:finding_best_train_test}}
  \label{table:baseline}
\end{table}
\subsubsection{Motion model analysis}
\label{sec:motion_model}
We analyze our architecture by exploring different model variations. The
original LSTM, as applied to speech processing, is
based on fully connected layers \cite{Graves14ICML, Graves13, Fernandez08}. In videos, such LSTM with fully connected layers,
applied after a
feature extraction phase (after the first convolutional layers), would
allow to model global motion of the object. We conjecture that in our application, an LSTM with
 convolutions
would allow to model local deformations of the object caused
by the motion. We experimented with different  LSTM architectures by varying the parameters
of convolutional-based LSTM and/or adding fully-connected layers to a fully-connected LSTM. Table \ref{table:model_analysis} compares the different architectures. 
\begin{table}[h!]
  
  \vspace*{0.22cm}
  \begin{tabular}{ | p{4cm}| c | c  |}
   \hline Model & Accuracy & Standard Deviation \\
\toprule
\multicolumn{3}{c}{Single frame dataset.} \\
   \hline  Baseline model  trained on single image.
   3 conv + 3 pools $\rightarrow$ FC $\rightarrow$ dropout $\rightarrow$ FC
    $\rightarrow$ dropout& 82.02 & 1.96 \\
    \hline
    \multicolumn{3}{c}{Short time-frame sequence dataset} \\
   \hline 3 conv + 3 pools with 128-dim FC LSTM & 75 & 3.14  \\
   \hline 3 conv + 3 pools $\rightarrow$ 512-dim FC $\rightarrow$
   128-dim FC LSTM & 79.27 & 3.3  \\
   \hline 3 conv + 3 pools  $\rightarrow$
   conv LSTM (filter size = 3, depth = 128) & 81.12 & 1.75  \\
   \hline 3 conv + 3 pools  $\rightarrow$
   conv LSTM (filter size = 5, depth = 256) where result of LSTM layers is summed & 82.24 & 2.49 \\
   \hline 3 conv + 3 pools  $\rightarrow$
   conv LSTM (filter size = 5, depth = 256) where result of LSTM layers is
concatenated & \textbf{82.59} & 1.78 \\

     \hline
   \end{tabular}
   \caption{ Performance of LSTM motion models on the Washington object dataset.
All models except the baseline were trained on short time-frame sequences. The following notation
was used when describing the network's architecture: conv - convolutional layer, FC - fully
connected layer.}
\label{table:model_analysis}
\end{table}
\par
\begin{figure}[h!]
  
  \centering
  \includegraphics[width=0.47\textwidth]{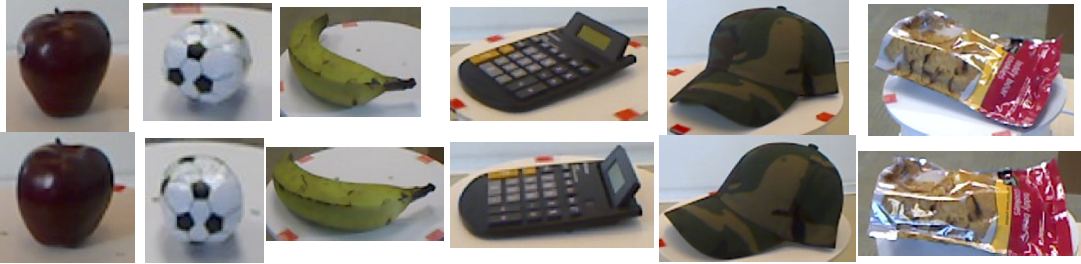}
  \caption{Example short time-frame sequences for the Washington-RGBD object dataset.}
  \label{fig:dataset_17}
\end{figure}

 In these experiments we use sequences which are more
realistic i.e. obtained in less than a second apart.
Thus, the dataset was created by mapping frame $t$  to a sequence $\{t - 17, t\}$ and is referred to as ``short time-frame'', Fig.~\ref{fig:dataset_17}.
Such mapping allows frames to be distinct enough to contain useful motion information, but
not too far apart as to be useful in a real world application (two frames in less than a second apart).
The results in Table \ref{table:model_analysis} suggest that LSTMs implemented as fully-connected
layers do not perform well and have high deviation. This might be due to overfitting since
the model can learn non-existing global patterns which are irrelevant to the object's identity.
The architectures based on convolutional LSTM gates perform significantly better and also have lower
standard deviation. Increasing the filter size of convolution and
increasing the depth of the channels improves the performance, as well. Although increasing filter size and depth
 significantly increases the computational cost, our best motion model runs at 0.8
 seconds per two-frame sequence, which is 4 times faster than the multi-frame baseline when
  averaged per image. We also experimented by stacking multiple LSTM layers
one of top of each other, but the accuracy was significantly lower than the models
presented in this section. We note that the results summarized in Table~\ref{table:baseline} (including for short-time frame sequences) are obtained by bidirectional testing, whereas Table~\ref{table:model_analysis} reports the unidirectional counterparts.

\subsubsection{Wide viewpoint sequences}
\begin{figure}[h!]
  \centering
  \includegraphics[width=0.4\textwidth]{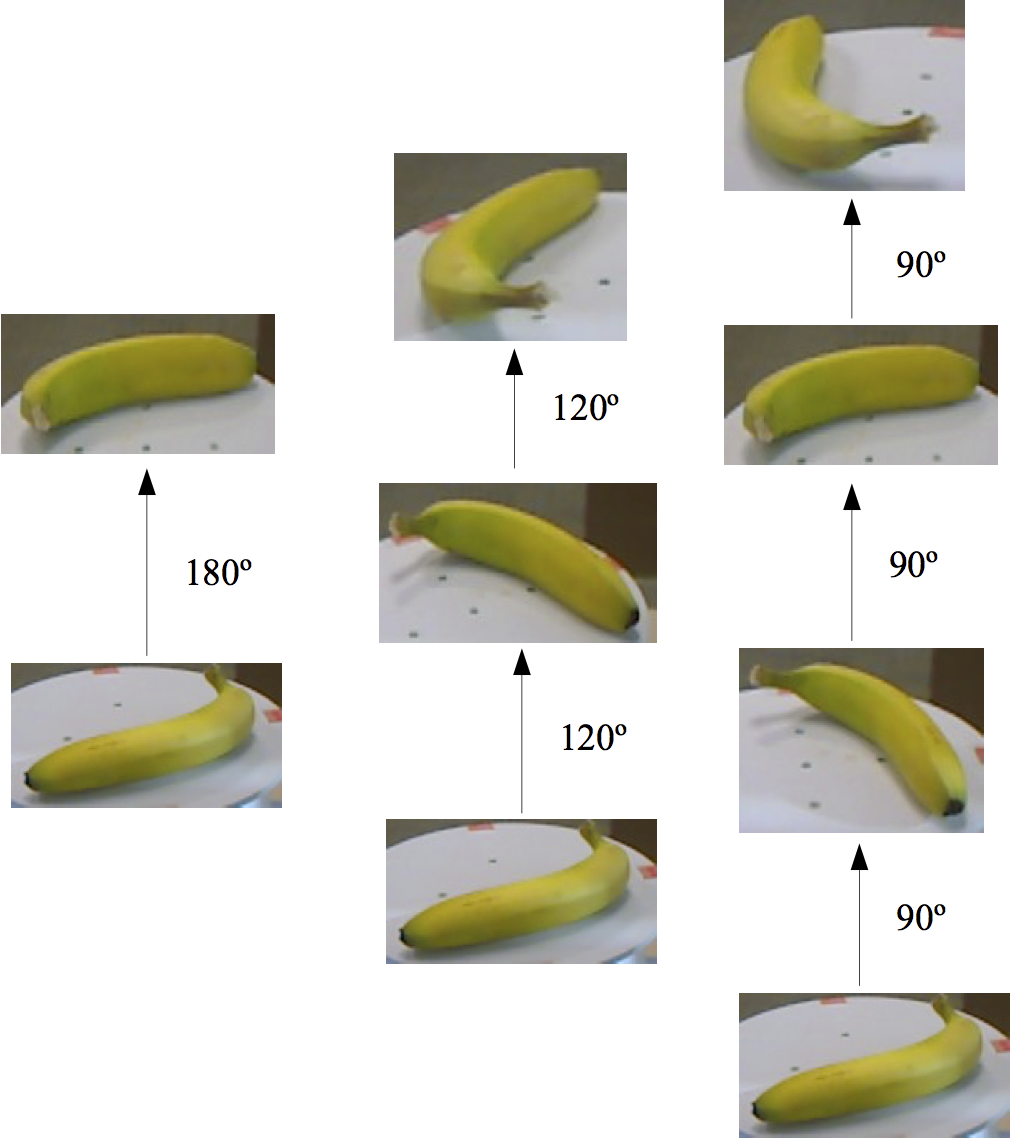}
  \caption{From left to right: example of the sequences generated for Washington dataset
with wide viewpoints for  $n=2, 3, 4$ frames. The frames are taken at large degree to
maximize viewpoint change.}
\label{fig:ideal_dataset}
\end{figure}

To see if our best motion model is capable of generalizing to more frames we created
wide viewpoint sequence datasets. Such sequences were created to take advantage of varying viewpoints of the objects.
The data was generated by taking frames to form a sequence if they are as far apart
as possible in terms of viewing angle. For example, for $n=2$
a sequence of two frames consists of the original first frame and the frame which
corresponds to the one where the object is rotated at 180 degrees.
Similarly, when $n>2$ we create a sequence of $n$ frames which have the
object rotated at $\frac{180}{n}$ degrees. An example of wide viewpoint sequence of banana for $n=2, 3, 4$ is
given in Fig. \ref{fig:ideal_dataset}.

\par Table \ref{table:baseline} shows the results of the motion model which was applied to wide viewpoint datasets. We can observe that using motion models is beneficial, even for as few as one additional frame (2-frame model). Furthermore, we see that adding more frames improves performance, but for the wide-viewpont model the accuracy levels off at 4 frames. This is as expected since all viewpoints are sampled on a sphere and at some point sampled frames would coincide with existing one in the training set, thus not adding any additional information. 

\subsubsection{Bidirectional models}
\label{sub:finding_best_train_test}
Bidirectional LSTM models \cite{schuster1997bidirectional}, where there is LSTM layer
for sequence in each of the directions, was found to be superior to unidirectional
models \cite{schuster1997bidirectional, Fernandez08, graves2005framewise}. We
found that such models are indeed superior for the classification task from videos.
Videos, unlike audio signals, are semantically sound regardless of the direction they
are played e.g. the video played backwards can still convey the same meaning. Thus, we opt to
investigate if it is possible to take advantage of this property at the \emph{data level}. To do so
we trained our model using dataset where sequences were passed in both directions
during training and/or testing. During training this effectively doubles the size of the
dataset while for testing we classify a sequence in each of the directions and average
the result.
\begin{table}[h!]
\vspace*{0.22cm}
  \begin{tabular}{ | p{4cm}| c | c  |}
   \hline Model  & Accuracy & Standard Deviation \\
   \toprule
\multicolumn{3}{c}{Short time-frame sequence dataset.} \\
   \hline Unidirectional train and test & 82.59 & 1.78  \\
   \hline Bidirectional training / unidirectional testing & 82.61 & 2.29 \\
   \hline Bidirectional training and testing & 82.68 & 2.27 \\
    \hline Bidirectional testing / unidirectional training & \textbf{82.74} & 1.76  \\
   \hline
   \end{tabular}
   \caption{Performance of the motion model with bidirectional training/testing}
 \label{table:bidirectional}
\end{table}
\par Results in Table \ref{table:bidirectional} suggest that bidirectional
training has almost no effect besides increasing the deviation. This result goes in hand
 with the
architecture of the network - bidirectional LSTM allows to pass information
in both directions, thus explicitly training it in this way is redundant. Although
bidirectional test is worse for 2-frame wide viewpoint dataset it increases the
accuracy for every other experiment.
\par On wide viewpoints datasets bidirectional testing showed a rather significant improvement over unidirectional one. For 3-frame dataset the accuracy increased from by more than $0.3 \%$ to $84.28 \%$ while for
$4$-frame dataset the increase was $0.45 \%$. On 2-frame datasets such strategy improved the accuracy on short time sequence dataset by $0.13 \%$, but decreased it for wide viewpoint one by $0.11 \%$.
\subsubsection{Pooling baseline}
\par We also compared to the baseline which uses all frames from the sequences available during
training. When given a sequence of frames during testing our baseline model would take an average
of the class probability distribution across all frames in a sequence. We also experimented
by taking the maximum instead of multi-frame average pooling, but the accuracy was always lower.

\subsection{Evaluation on Washington Scenes dataset}
\begin{figure*}
  \centering
  \includegraphics[width=1\textwidth]{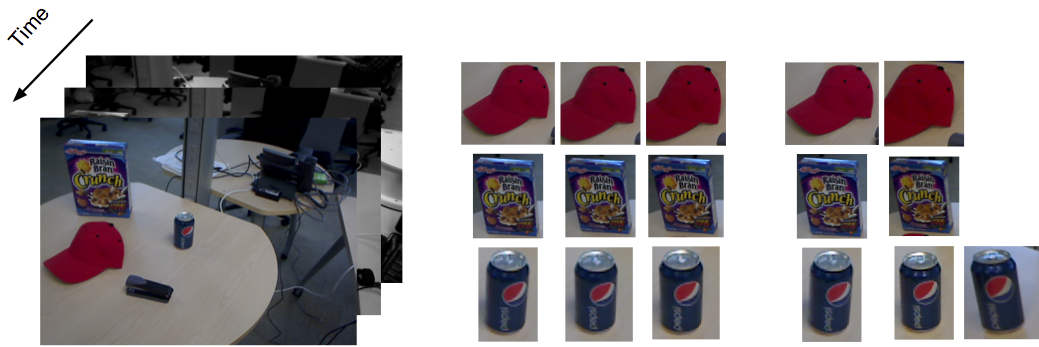}
  \caption{Example of a sequence and object crops on Washington Scenes dataset \cite{lai2011large}. Left: video frames. Middle: crops of objects at 3 frames apart. Right:
crops at 10 frames apart.}
\label{fig:extra_dataset}
\end{figure*}

\begin{figure}[h!]
  \centering
  \includegraphics[width=0.5\textwidth]{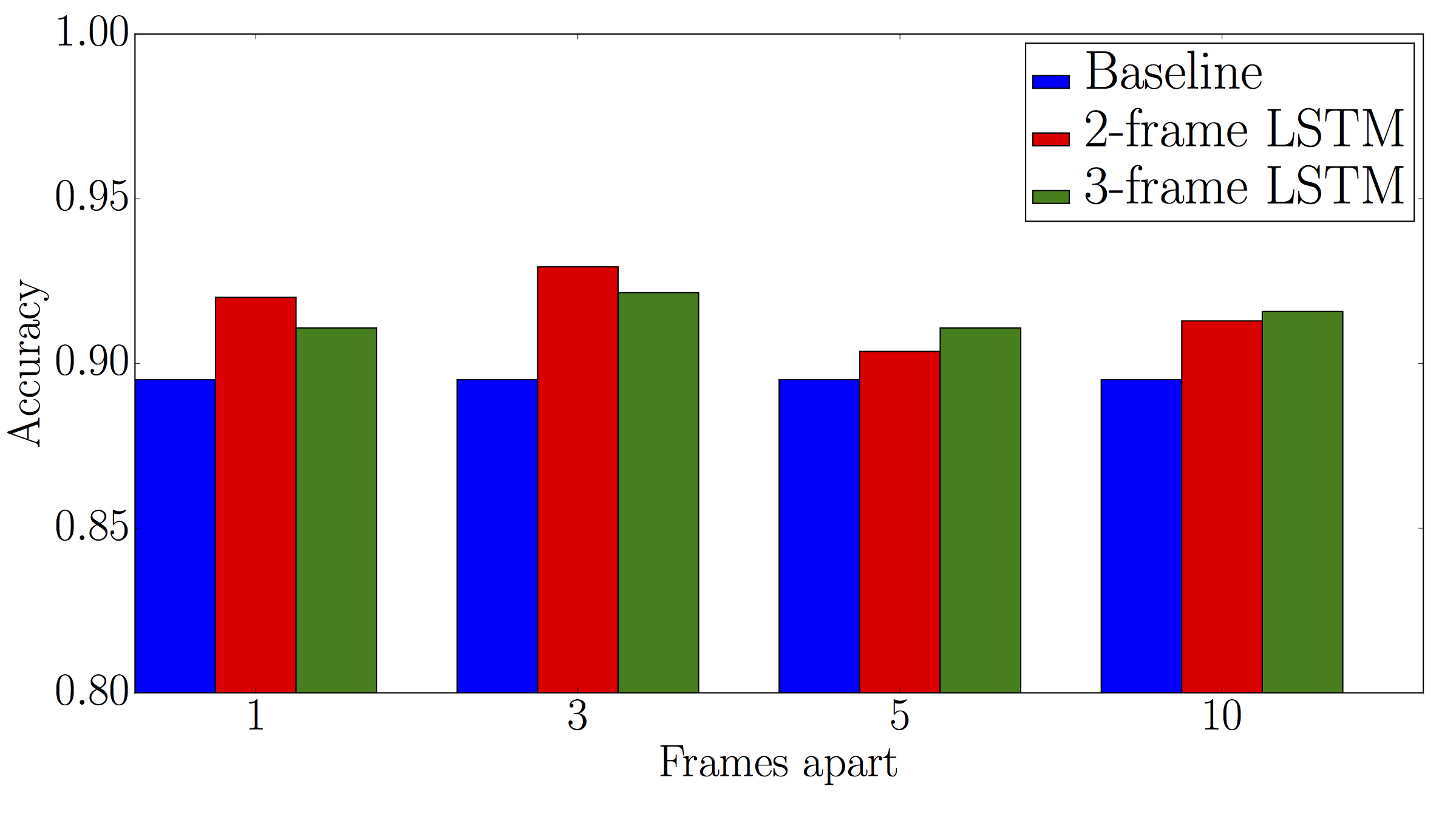}
  \caption{Results on the Washington Scenes dataset. The three columns per
  experiment denote results on the baseline model
on still images, 2-frame LSTM and 3-frame LSTM respectively. x-axis denotes how
far were the frames in the sequences. For example, 5 denotes sequences $\{t-10, t-5, t\}$.}
\label{fig:extra_dataset_image}
\end{figure}

To see how well our method works on more realistic videos we performed experiments
on Washington Scenes dataset \cite{lai2011large} (see Fig. \ref{fig:extra_dataset}). The dataset consists of five
videos of the office containing the following objects: soda can, bowl, cap, coffee mug, cereal box
and flashlight. We cropped objects throughout videos and used one independent video ``table\_small'' for testing
and the rest for training. Because the test set did not contain a single image of the flashlight
we removed the first 200 frames of the flashlight and added them to the test set, without any
overlap. We evaluated our model by taking 1, 3, 5, and 10 prior frames to form sequences. For example, for 10 and for a 3-frame sequence, we compose a sequence of frames $\{t - 20, t - 10, t \}$. If the frames did not exist or
the objects were not visible at a particular frame, the original crop was used instead, although other solutions are also possible.
Fig. \ref{fig:extra_dataset_image} contains an example of a video frame and related object crops.

Results are reported in Fig. \ref{fig:extra_dataset_image}. These experiments also confirm
 that our motion model is beneficial as it is capable of using
 motion information from sequences. As images in sequences become more far apart, from 1 to 3 frames, our motion model improves as the latter
 sequences contain more motion information. 
We note that for this dataset, we have used the same network architecture and learning hyper-parameters as for Washington Object dataset, so these results are an indicator of the universality and the generalization capabilities of our proposed model.
It can also be seen from Fig.~\ref{fig:extra_dataset_image}, that the accuracies
decrease for 5 and 10 frame dataset a bit probably due to bad quality of the data, which is due to the fact that the object goes out-of-view in the original scene videos: more than 50 \% and
80 \% of sequences, respectively, had  to use shorter sequences, as the object was not visible (i.e. the corresponding frames were
missing). Thus we observe, that in practical applications, learning needs to be done from frames that are not too far apart; although in principle multiple frames are capable of providing extra information, in practice, these frames may often not be available and thus are not useful.
This also enforces our view that learning from short videos is the most cost-effective for object perception.

\subsection{Implementation details}
\subsubsection{Baseline model parameters} 
\label{sub:baseline_section_details}
The convolutional layers parameters are: stride 2, filter size 5,
depth 64 for the first convolutional layer; stride 1, filter size 3, depth 128 for the second and  the third. Each pooling layer has a stride of 2 in $x$ and $y$ direction. Each of the fully connected layers
has a dimension of 4096; dropout probability is set to 0.5. In total, our baseline model has 77 million parameters.

\subsubsection{Initialization}
Deep CNNs allow to train powerful classifiers, but require significant amount of
data to do so. To overcome this, CNNs are often initialized from networks
trained on large datasets like ImageNet\cite{Deng09}, e.g. \cite{Girshick14}. In all of our models
we initialize the first three convolutional layers by pre-training a network, as in \cite{Krizhevsky12}, on ImageNet.
\subsubsection{Learning rates} All models using baseline architecture were
trained with the learning rate of $0.0001$. For motion models we used $0.001$.
\subsubsection{Runtime}
We timed our models to see how fast can they perform recognition. Baseline model: 0.1 s/image;
two-frame motion model: 0.87 s/image. The timings are reported using single NVidia K40 GPU.

\section{Conclusion}
\label{sec:conclusion}
This paper proposes using motion information to improve object recognition for robotics perception. Here
we used motion information from very short video sequences (e.g. 2-4 frames) to improve object recognition. We designed a recurrent
neural network, based on convolutional-only LSTM, capable of utilizing motion information
to guide the recognition task. We showed that bidirectional LSTM with unidirectional training
and bidirectional testing is the best model for this purpose. Our experiments on
wide viewpoint and natural videos  showed that
such motion model is capable of generalizing when more frames are available and is able to
outperform baseline non-recurrent convolutional network.

\section{Acknowledgements}
The authors would like to thank Alex Krizhevsky for his help and comments, and the Google Brain team for providing inspiring
environment and computational resources to carry out experiments.
\bibliography{IEEEabrv,egbib}
\bibliographystyle{IEEEtran}
\end{document}


\maketitle
\thispagestyle{empty}
\pagestyle{empty}

  \section{Network Architectures}

  \subsection{Baseline model}
  \label{sub:Baseline model}
\begin{figure*}[H]
    \label{fig:baseline}
    \centering
    \includegraphics[width=1\textwidth,height=1\textheight,keepaspectratio]{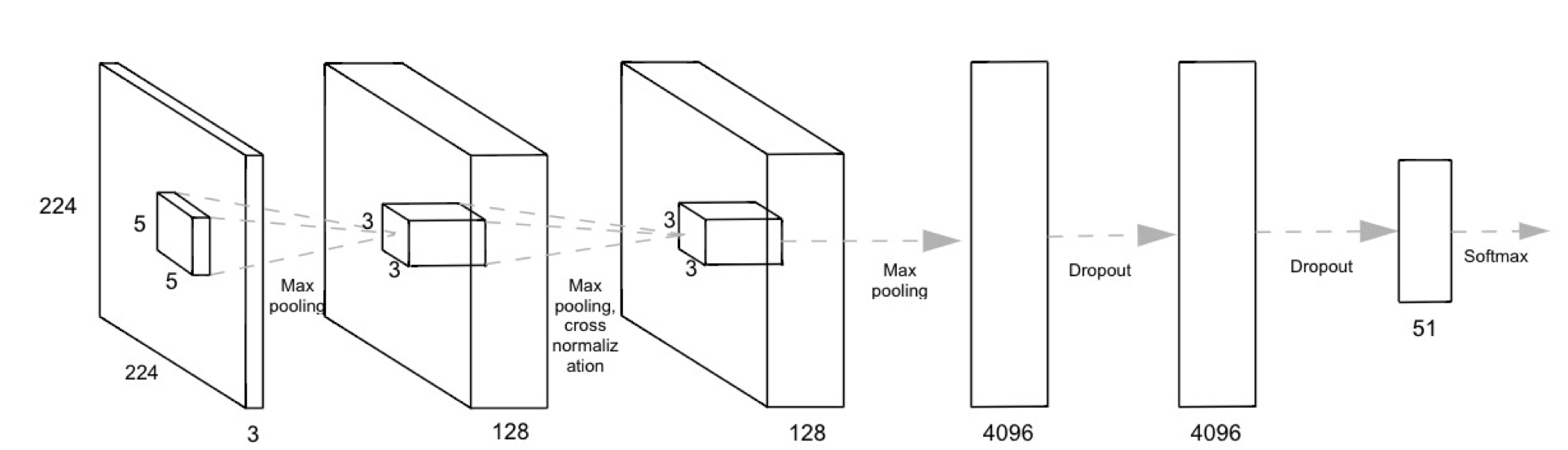}
    \caption{Baseline model architecture.}
\end{figure*}

  Convolutional network architecture, referred in text as baseline, is shown in Figure \ref{fig:baseline}.
We build our baseline architecture by stacking three convolutional layers on top of each
other followed by two fully connected layers. Each convolutional layer is
followed by the pooling with local contrast normalization
 after second pooling layer. Both fully connected layers are followed by the
 dropout layer each. Result of the second fully connected layer is fed into
 another fully connected layer whose response is used as an input to the
 softmax function which outputs a probability distribution of objects being
 in the image.

\subsection{Motion model}
\label{sub:Motion model}
\begin{figure*}
  \label{fig:motion}
  \centering
  \includegraphics[width=\textwidth,height=\textheight,keepaspectratio]{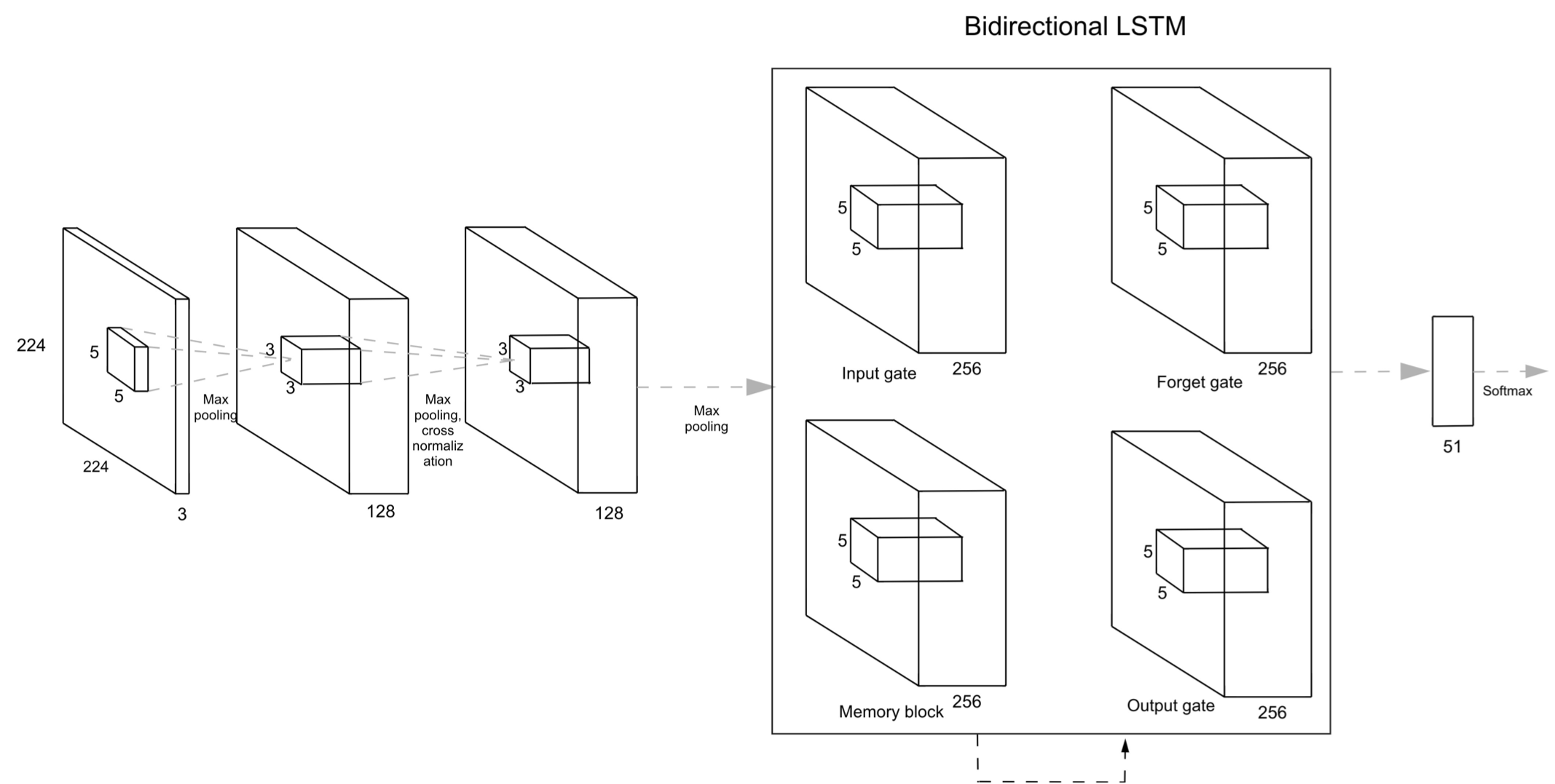}
  \caption{Motion model architecture.}
\end{figure*}

The architecture of the motion model is shown in Figure \ref{fig:motion}. The first three layers in the motion model are the same as in the baseline. After the third pooling layer, we place forward and backward LSTM layers where each gate is a convolution. Each convolutional gate has the same parameters: stride 1, filter size 5, depth 256. Result of the both LSTM layers is concatenated and fed into fully connected layers with softmax function on top.

